 \documentclass[10pt,a4paper]{article}
\usepackage[T1]{fontenc}
\usepackage{inputenc}
\usepackage[english]{babel}
\usepackage[nohead,includefoot,margin=2cm]{geometry}
\usepackage[compact,raggedright,small]{titlesec}
\usepackage[affil-it]{authblk}
\usepackage[runin]{abstract}
\usepackage{multicol}
\usepackage{graphicx}

\newcommand*{\pacsname}{PACS numbers}

\addto{\Affilfont}{\small}

\title{\bfseries\large Data Assimilation by Artificial Neural Networks for an Atmospheric 
        General Circulation Model: Conventional Observation}
\author[]{Ros\^angela S. Cintra}
\author[]{Haroldo F. de Campos Velho}
\affil[]{Brazilian National Institute of Space Research, S. Jos\'e dos Campos, SP, Brazil}

\begin{document}
\maketitle

\begin{abstract}
This paper presents an approach for employing an artificial neural network (NN) 
to emulate an ensemble Kalman filter (EnKF) as a method of data assimilation. 
The assimilation methods are tested in the Simplified Parameterizations 
PrimitivE-Equation Dynamics (SPEEDY) model, an atmospheric general circulation 
model (AGCM), using synthetic observational data simulating localization of balloon soundings. 
For the data assimilation scheme, a supervised NN, the multilayer perceptron 
(MLP-NN), is applied. The MLP-NN is able to emulate the analysis from the 
local ensemble transform Kalman filter (LETKF). After the training process, 
the method using the MLP-NN is seen as a function of data assimilation. 
The NN was trained with data from first three months of 1982, 1983, and 1984. 
A hind-casting experiment for the  1985 data assimilation cycle using MLP-NN 
was performed with synthetic observations for January 1985. 
The numerical results demonstrate the effectiveness of the NN technique 
for atmospheric data assimilation. The results of the NN analyses are 
very close to the results from the LETKF analyses, the differences of 
the monthly average of absolute temperature analyses is of order 
$10^{-2}$. The simulations show that the major advantage of using the 
MLP-NN is better computational performance, since the analyses have similar 
quality.  The CPU-time cycle assimilation with MLP-NN is $90$ times faster than 
cycle assimilation with LETKF for the numerical experiment.
\end{abstract}

\begin{multicols}{2}
 
\section{Introduction}  
\noindent
For operating systems in weather forecasting, one of the challenges is 
to obtain the most appropriate initial conditions to ensure the best 
prediction from a physical-mathematical model that represents the evolution 
of a physical system. Performing a smooth melding of data from observations 
and model predictions carries out a set of procedures to determine the best 
initial condition. Atmospheric observed data are used to create meteorological 
fields over some spatial and/or temporal domain. Data assimilation occurs 
when the observations and the dynamic model are combined. 

The analysis for the time evolution of the atmospheric flow is based on 
observational data and a model of the physical system, with some background 
information on initial condition. The analysis is useful in itself as a description 
of the physical system, but it can be used as an initial state for the further 
time evolution of the system \cite{Holm08}.
 
Several techniques have been developed to identify the initial condition for numerical 
weather prediction (NWP). These techniques are applied in models of atmospheric and 
oceanic dynamics, environmental and hydrological prediction, and ionosphere dynamics. 
The Kalman filter (KF) \cite{Kalman60} is one strategy to estimate an appropriate 
initial condition. Another strategy is to find a probability density function associated 
with the initial condition, characterizing the Bayesian approaches  
\cite{Daley91,Lorenc86}. The ensemble Kalman filter (EnKF)  \cite{Evensen94, 
Houtekamer98} and particle filter (PF)  \cite{Doucet02, Andrieu02}, 
are Bayesian techniques.

These modern techniques represent a computational challenge, even with the use of 
parallel computing with thousands of processors. Indeed, the challenge is the 
computational cost because we are moving to higher resolution models and an 
exponential growth in the amount of observational data. The algorithms are constantly 
updated to improve their performance. One example is the version of the EnKF restricted 
to small areas (local): the local ensemble Kalman filter (LEKF) \cite{Ott04}. 
The computational challenge to the techniques of data assimilation,  
lies in the size of matrices involved in operational NWP models, 
currently running under order of millions of equations 
(equivalent to full matrix elements of the order of $10^{12}$).  

The application of Artificial Neural Networks (NN) was suggested as a possible 
technique for data assimilation by \cite{HsiehTang98}, \cite{Liaqat03}, and \cite{Tang01}. 
However, \cite{Nowosad00} (see also \cite{Vijay02, CamposVelho02}) 
employed the first implementation of an NN as an approach for data 
assi\-mi\-la\-tion, with independent development from the cited authors; 
they used an NN over all spatial domain. Later, this approach was improved by 
\cite{Harter05} and \cite{Harter08}, they analyzed the performance of two feed-forward 
NN (multilayer perceptron and radial basis function), and two recurrent NN 
(Elman and Jordan) (see \cite{Haykin01, Haykin07}). \cite{Harter08} 
introduced a modification on the NN application, in which the analysis 
was obtained at each grid point, instead of at all points of the domain. 
The modification greatly reduced the dimension of the the computational processing.  
Continuing the investigations, \cite{Furtado08} evaluated the 
performance of an NN to emulate the particle filter and the variational method for 
data assimilation applied to Lorenz chaotic system. The NN technique was 
successful for all experiments.

The use of the NN in data assimilation approach does not address error estimation. 
In these experiments, the NN were applied to mimic other data assimilation methods. 
The main advantage to using NN is the speed-up of the data assimilation process. 
Methods using NN have shown consistent results with implementation in the 
simple and low-order models.

This paper presents the approach based on using NN to emulate 
an EnKF version as a method of data assimilation. 
The NN are applied to a nonlinear dynamical system, 
an atmospheric general circulation model (AGCM). The AGCM used, is the 
Simplified Parameterizations PrimitivE-Equation Dynamics (SPEEDY) model. 
The method is tested with synthetic conventional data, simulating measurements
from surface stations (data at each 6 hours on a day) and upper-air soundings (data at 
each 12 hours on a day). The application of NN produces a significant reduction for 
the computational effort compared to LETKF. The goal to use the NN approach 
is to achieve a better computational performance with similar quality for the 
prediction, i.e., an computational efficient process of atmospheric data 
assimilation (the analysis).

The technique uses NN to implement the function:
\begin{equation}
           x^a = F_{NN}(y^o, x^f)
\end{equation}
where $F_{NN}$ is the data assimilation process, $y^o$ are the observations, 
$x^f$ is a model forecast (simulated), often called the first guess, 
and $x^a$ is the analysis field with innovation that represents the 
correction to the model. 

Generally, the observational data used in operational data assimilation are 
conventional data and satellite data. The conventional data include surface 
observations and upper-air observations, such as balloon soundings. 
Global operational satellite data are taken and processed in real time to 
all Earth surface. Though small in number, the conventional data are very 
important in the meteorological data assimilation. 
Then, we apply that synthetic data type in our experiment.

The experiment was conducted using the SPEEDY model 
\cite{Bourke74, Held94}, which is a 3D global atmospheric model, 
with simplified physics parameterization by \cite{Molteni03}. 
The spatial resolution considered is T30L7 for the spectral method. 
The grid of synthetic observations seeks to reproduce the stations 
of World Meteorological Organization (WMO) of rawinsonde observations. 
This spatial grid points are employed in this numerical experiment. 
Here, a set of NN \textit{multilayer perceptron} (MLP) 
[see \cite{Haykin01}] is employed to emulate the LETKF. 
The LETKF technique was used as the reference analysis. 
More information about LETKF can be obtained from 
\cite{Bishop01, Hunt07} and \cite{Miyoshi07}.

Our paper shows that the analysis computed by the NN has the 
same quality as the analysis produced by LETKF, which was analyzed 
by expert meteorologists, see \cite{Miyoshi05} and \cite{Kalnay03}.

%
\vspace*{5mm}
\section{Methodology}

\subsection{Artificial Neural Network (NN)}
An NN is a computational system with parallel and distributed processing 
that has the ability to learn and store experimental knowledge. An NN is 
composed of simple processing units that compute certain mathematical functions 
(usually nonlinear). An NN consists of interconnected artificial neurons or nodes, 
which are inspired by biological neurons and their behaviour. 
The neurons are connected to others to form a NN, which is used to model 
relationships between artificial neurons.
\begin{figure*}[t]
\vspace*{2mm}
\begin{center}
\includegraphics[width=9cm]{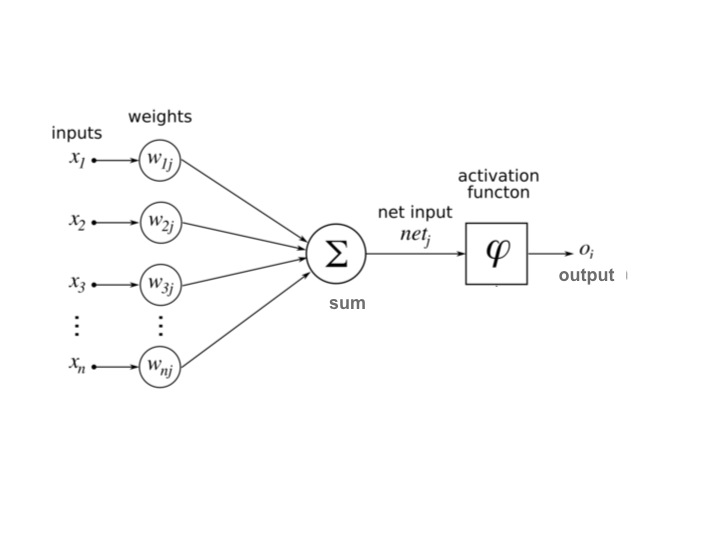}
\end{center}
\caption{Artificial Neural Neuron components.}
\label{fig:fig1} 
\end{figure*}

Each artificial neuron is constituted by one or more inputs and one output. 
The neuron process is nonlinear, parallel, local, and adaptable. Each neuron 
has a function to define outputs, associated with a learning rule. The neuron 
connection stores a nonlinear weighted sum, called a weight. In NN processing, 
the inputs are multiplied by weights; these results summarized then 
go through the activation function. This function activates or inhibits 
the next neuron. Mathematically, we can describe the 
$i^{th}$ input with the following form: 
\begin{equation} \label{rdelta}
\begin{array}{ll}
\mbox{input summation:} & u_{i} = \sum^p_{j=1}{w_{ij}x_j }  \\
\mbox{neuron output:}   & y_i   = \varphi(u_i)
\end{array}
\end{equation}
where $x_1,x_1,.\cdots,x_p$ are the inputs; $w_{i1},\cdots,w_{ip}$ are the 
synaptic weights; $u_i$ is the output of linear combination; $\varphi(\cdot)$ 
is the activation function, and $y_i$ is the $ith$ neuron output, 
$p$ is number of neurons (Fig. 1).

A feed-forward network, which processes in one direction from input to 
output, it has a layered structure. The first layer of an NN is called the 
input layer, the intermediary layers are called hidden layers, and the last layer 
is called the output layer. The number of layers and the quantity of neurons in each 
is determined by the nature of the problem. In most applications, a feed-forward NN 
with a single layer of hidden units is used with a sigmoid activation function, 
such as the hyperbolic tangent function for the units
\begin{eqnarray} \label{tgh}
    \mbox{hyperbolic tangent function}:  \nonumber \\ 
    \varphi(v)=\frac{1-\exp(-av)}{1+\exp(-av)}~.
    \label{tanh}
\end{eqnarray}

There is two distinct phases in using an NN: the training phase (learning process) and 
the run phase (activation or generalization). The training phase of the NN consists of 
an iterative process for adjusting the weights for the best performance of the NN 
in establishing the mapping of input and target vector pairs. The learning algorithm is 
the set of procedures for adjusting the weights. The single pass through the 
entire training set in one process in called an epoch. 
Testing of the verification set follows each epoch, 
the iterative process continues or stops after verification of defined criteria, which 
can be minimum error of mapping or number of epochs. 
Once the NN is trained (the process is stopped) , the weights 
are fixed, and the NN is ready to receive new inputs (different from training inputs)
for which it calculates the corresponding outputs. 
This phase is called the generalization. Each connection (after training) 
has an associated weight value that stores the knowledge represented in the experimental 
problem and considers the input received by each neuron of that NN.

Neural network designs or architectures  are dependent upon the learning strategy adopted 
[see Haykin \cite{Haykin01}]. 
The multilayer perceptron (MLP)  is the NN architecture used in this study; which
the interconnections between the inputs and the output layer have at least one 
intermediate layer of neurons, a hidden layer [\cite{Gardner98}, \cite{Haykin07}].
 
Neural Networks can solve nonlinear problems if nonlinear 
activation functions are used for the hidden and/or the output layers. The use of units 
with nonlinear activation functions generalizes the delta rule. Developed by 
\cite{Widrow60}, the delta rule is a version of the least mean square (LMS) method. 
For a given input vector $x$, the output vector $x^a_{NN}$ is compared to the target answer
$x^a_{ref}$. If the difference is near zero, no learning takes place; 
if the difference is not near zero, the weights are 
adjusted to reduce this difference. The purpose is to minimize the output errors by adjusting 
the NN weights, $w_{ij}$, using the delta rule algorithm, summarized as follows: 
 
\begin{enumerate}
\item[1.] Compute the error function $E(w_{ij})$, defining the distance between the
         target and the result: $E(w_{ij}) \equiv [x^a_{ref}-x^a_{NN}]^2$
\item[2.] Compute the gradient of the error function 
	 $\partial {E(w_{ij})}/\partial{w_{ij}} = \delta_j y_i$, defining which direction 
	 should move in weight space to reduce the error, and $\delta_j \equiv 
	 (x^a_{ref}-x^a_{NN})\varphi'(v)$.
\item[3.] Select the learning rate $\eta$ which specifies the step size taken in the 
         weight space of update equation;
\item[4.] Update  the weight for the epoch $k$: 
         $w_{ij}^k =w_{ij}^{k-1} +   \Delta w_{ij}$, where
         $\Delta w_{ij} = -\eta {E(w_{ij})}/\partial{w_{ij}}$.
         One epoch or training step is a set of update 
         weights for all training patterns.
\item[5.] Repeat step[4] until the NN error function, reach the required precision.
         This precision is a defined parameter that stops the iterative process. 
\end{enumerate}

The overall idea is to treat the NN as a function 
of the weights $w_{ij}$ (Eq.~\ref{rdelta}), instead of the inputs. 
The goal is to minimize the error between the actual 
output $y_i$ (or $x^a_{NN}$) and the target output ($d_i$) 
(or $x^a_{ref}$) of the training data. 
For each (input/output) training pair, the delta rule determines 
the direction you need to be adjusted to reduce the error. 
By taking short steps, we can find the best direction for the entire training.
To consider a set of (input and target) pairs of vectors 
$\{ (x_0, d_0),\  (x_1,d_2), \cdots, (x_N,d_N) \}^T$, where $N$ 
is the number of patterns (input elements), and 
one output vector $y=[ y_0,y_1,y_2,\cdots,y_N]^T$, 
a MLP performs a complex mapping $y=\varphi(w,x)$ 
parameterized by the synaptic weights $w$, and the functions 
$\varphi(\cdot)$ that  provide the activation for the neuron. 

The set of procedures to adjust the weights is the learning algorithm. 
\textit{back-propagation}, a well-known learning scheme, 
is generally used for the MLP training (it performs the delta rule 
(the algorithm above). 
Back-propagation training is a supervised learning., e.g. 
the adjustments to the weights are conducted by back propagating, or 
considering the difference between the NN calculated output 
and the target output (considered the supervisor).

\cite{Hsieh09}, and \cite{Haupt09} reviewed applications of NN in 
environmental science including in atmospheric sciences. They reviewed some NN 
concepts and some applications, they presented others estimation methods and its 
applications. The NN applications, generally, are on function approximation of 
modeling of nonlinear transfer functions, and pattern classifications.  
\cite{Gardner98} included brief introductions of MLP and the back-propagation 
algorithm; they showed a review for applications in the atmospheric sciences, 
looking at prediction of air-quality, surface ozone concentration, dioxide concentrations, 
severe weather, etc., and pattern classifications applications in remote sensing data to 
obtain distinction between clouds and ice or snow. Applications on classification of 
atmospheric circulation patterns, land cover and convergence lines from radar imagery, 
and classification of remote sensing data using NN, were also presented in 
Hsieh (2009) and Haupt et al. (2009). 
\vspace*{5mm}

%
%
\subsection{The SPEEDY Model}
\noindent
The SPEEDY computer code is an AGCM developed to study global-scale dynamics 
and to test new approaches for NWP. The dynamic variables for the primitive 
meteorological equations are integrated by the spectral method in the 
horizontal at each vertical level (see \cite{Bourke74, Held94}). 

The model has a simplified set of physical parameterization schemes that 
are similar to realistic weather forecasting numerical models. The goal of 
this model is to obtain computational efficiency while maintaining characteristics 
similar to the state-of-the-art AGCM with complex physics parameterization 
\cite{Miyoshi05}.
 
According to~\cite{Molteni03}, the SPEEDY model simulates the general structure 
of global atmospheric circulation fairly well, and some aspects of the systematic 
errors are similar to many errors in the operational AGCMs. The package is based 
on the physical parameterizations adopted in more complex schemes of the AGCM, 
such as convection (simplified diagram of mass flow), large-scale condensation, 
clouds, short-wave radiation (two spectral bands), long--wave radiation 
(four spectral waves), surface fluxes of momentum, energy (aerodynamic formula), 
and vertical diffusion. Details of the simplified physical parameterization 
scheme can be found in~\cite{Molteni03}. 

The boundary conditions of the SPEEDY model includes topographic height and 
land-sea mask, which are constant. Sea surface temperature (SST), sea ice 
fraction, surface temperature in the top soil layer, moisture in the top soil layer, 
the root-zone layer, snow depth, all of which are specified by monthly means. 
 Bare-surface albedo, and fraction of land-surface covered by vegetation,  
are specified by annual-mean fields. The lower boundary conditions 
such as SST are obtained from the ECMWF's reanalysis in the period 1981-90. 
The incoming solar radiation flux and the boundary conditions (SST, etc.), 
are updated daily. The SPEEDY model is a hydrostatic model in sigma 
coordinates, \cite{Bourke74} also describes the vorticity-divergence 
transformation scheme.

The SPEEDY model is global with spectral resolution T30L7 
(horizontal truncation of 30 numbers of waves and seven levels), 
corresponding to a regular grid with 96 zonal points (longitude), 
48 meridian points (latitude), and 7 vertical pressures levels 
(100, 200, 300, 500, 700, 850, and 925 hPa). 
The prognostic variables for the model input and output are the 
absolute temperature ($T$), surface pressure ($p_s$),
zonal wind component ($u$), meridional wind component ($v$), 
and an additional variable and specific humidity ($q$). 
\vspace*{5mm}

%
%
\subsection{Brief Description on Local Ensemble Transform Kalman Filter}
\label{letkf}
\noindent
The analysis is the best estimate of the state of the system based on the optimizing 
criteria and error estimates. The probabilistic state space formulation and the 
requirement for updating information when new observations are encountered are 
ideally suited to the Bayesian approach. The Bayesian approach is a set of efficient 
and flexible Monte Carlo methods for solving the optimal filtering problem. 
Here, one attempts to construct the posterior probability density function 
(pdf) of the state using all available information, including the set of 
received observations. Since this pdf embodies all available statistical information, 
it may be considered as a complete solution to the estimation problem. 

In the field of data assimilation, there are only few contributions 
in sequential estimation (EnKF or PF filters). The EnKF was first proposed by 
 \cite{Evensen94} and was developed by \cite{Burgers98} and \cite{Evensen03}. 
It is related to particle filters in the context that a particle is identified
as an ensemble member. EnKF is a sequential method, which means that the model 
is integrated forward in time and whenever observations are available; these 
EnKF results are used to reinitialise the model before the integration continues. 
The EnKF originated as a version of the Extended Kalman Filter (EKF) 
[\cite{Kalman61}].
The classical KF method by \cite{Kalman60} is optimal in the sense of 
minimizing the variance only for linear systems and Gaussian statistics. 
Analysis perturbations are added to run the ensemble forecasts.
\cite{Miyoshi07} added Gaussian white noise to run the same forecast 
for each member of the ensemble in LETKF. 
The EnKF is a Monte Carlo integration that governs the evolution of the pdf,
which describes the \textit{a priori} state, the forecast and error statistics. 
In the analysis step, each ensemble member is updated according 
to the KF scheme and replaces the covariance matrix by the sampled covariance 
computed from the ensemble forecasts.

\cite{Houtekamer98} first applied the EnKF to an atmospheric system. They applied 
an ensemble of model states to represent the statistical model error. 
The scheme of analysis acts directly on the ensemble of model states when observations 
are assimilated. The ensemble of analysis is obtained by assimilation for each member 
of the reference model. Several methods have been developed to represent the modeling 
error covariance matrix for the analysis applying the EnKF approach; the local ensemble 
transform Kalman filter (LETKF) is one of them.  
\cite{Hunt07} proposed the LETKF scheme as an efficient upgrade of the local 
ensemble Kalman filter (LEKF). The LEKF algorithm creates a close 
relationship between local dimensionality, error growth, and skill of the ensemble 
to capture the space of forecast uncertainties, formulated with the EnKF scheme 
(e.g., \cite{Whitaker02}). 
In addition, \cite{Kalnay03} describes the theoretical foundation of the operational 
practice of using small ensembles, for predicting the evolution of uncertainties 
in high-dimension operational NWP models. 

The LETKF scheme is a model-independent algorithm to estimate the state 
of a large spatial temporal chaotic system \cite{Ott04}. The term ''local'' 
refers to an important feature of the scheme: it solves the Kalman filter 
equations locally in model grid space. a kind of ensemble square root filtering 
\cite{Miyoshi05},\cite{Whitaker02}, in which 
the analysis ensemble members are constructed by a linear combination of 
the forecast ensemble members. The ensemble transform matrix, composed 
of the weights of the linear combination, is computed for each local subset 
of the  state vector independently, which allows essentially parallel 
computations. The local subset depends on the error covariance localization 
\cite{Miyoshi12}. Typically a local subset of the state vector contains all 
variables at a grid point. The LETKF scheme first separates a global grid 
vector into local patch vectors with observations. 

The basic idea of LETKF is to perform analysis at each grid point 
simultaneously using the state variables and all observations in 
the region centred at given grid point.The local strategy separates 
groups of neighbouring observations around a central point for a given 
region of the grid model. Each grid point has a local patch; the number of 
local vectors is the same as the number of global grid points \cite{Miyoshi07}.
 
The algorithm of EnKF follows the sequential assimilation steps of classical 
Kalman filter, but it calculates the error covariance matrices as described bellow:

Each member of the ensemble gets its forecast 
\[ \{x^f_{n-1}\}^{(i)} \ : \ i=1,2,3, \cdots ,k, \]
where $k$ is the total members at time $t_n $, 
to estimate the state vector $\bar{x}^f$ of the reference model. 
The ensemble is used to calculate the mean of forecasting $(\bar{x}^f)$
\begin{equation}
\bar{x}^f \equiv k^{-1} \sum^k_{i=1}\{x^f\}^{(i)}~.
\end{equation}			      			                         
Therefore, the model error covariance matrix is:
\begin{equation}
P^f =(k-1)^{-1}\sum^k_{i=1} {( \{ x^f \}^{(i)} - \bar{x}^f)(\{x^f\}^{(i)} 
        - \bar{x}^f)^T}.
\end{equation}
The analysis must determine a state estimate and the covariance error matrix  
$P^a$, but also an ensemble with the appropriate sample analyses mean
$\{x^a_{n-1}\}^{(i)} \ : \ i=1,2,3, \cdots ,k, $   
\begin{eqnarray}
\bar{x}^a \equiv k^{-1} \sum^k_{i=1}\{x^a\}^{(i)}~ \\
P^a =(k-1)^{-1}\sum^k_{i=1} {( \{ x^a \}^{(i)} - \bar{x}^a)(\{x^a\}^{(i)} 
        - \bar{x}^a)^T}.
\end{eqnarray}

Important properties of the LETKF include the following:  
\begin{itemize}
\item Solve the estimation problem independently at each local region; 
\item Update the estimate of the current state and also its uncertainty; 
\item Assimilate all data at once;
\item Can deal with observations from the nonlinear functions of the state vector; 
\item Interpolate observations in time as well as in space (full 4D assimilation scheme); 
\item Set local region size and ensemble size (free parameters only); 
\item Be model independent (no adjoints!);
\item Can estimate model parameters errors as well as initial conditions.
\end{itemize}

The LETKF code has been applied to a low-dimensional 
AGCM SPEEDY model \cite{Miyoshi05}, a realistic model 
according \cite{Szunyogh07}. The LETKF scheme was also 
applied in: the AGCM for the Earth Simulator by \cite{Miyoshi07} 
and the Japan Meteorological Agency operational global and 
mesoscale models by\cite{Miyoshi10}; 
The Regional Ocean modeling System  by \cite{Shchepetkin05}; 
the global ocean model known as the Geophysical Fluid Dynamics Laboratory (GFDL) 
by \cite{Penny11}; and GFDL Mars AGCM  by \cite{Greybush05}.

%
\vspace*{5mm}
\section{MLP-NN in assimilation for SPEEDY model}
\label{MLPDA}

The NN configuration for this experiment is a set of multilayer perceptrons, 
hereafter referred to as MLP-NN, with the following characteristics:  
\begin{itemize} 
\item  [1.] Two input nodes, one node for the meteorological observation vector and 
      the other for the 6-hours forecast model vector. The vectors values represent individual 
      grid points for a single variable with a correspondent observation value.
\item  [2.] One output node for the analysis vector results. In the training algorithm, 
      the MLP-NN computes the output and compared it with the analysis vector of LETKF 
      results (the target data, but not a node for the NN). The vectors represent 
      individual analysis values for one grid point. 
\item  [3.] One hidden layer with eleven neurons. 
\item  [4.] The hyperbolic tangent (Eq. \ref{tgh}) as the activation function 
      (to guarantee the nonlinearity for results).
\item  [5.] Learning rate $\eta$ is $0.001$
\item  [6.] Training stops when the error reaches $10^{-5}$ or after 5000 epochs,
       which criterion first occurs.
\end{itemize}
 
On the present paper, the NN configuration (number of layers, nodes per layer, 
activation function, and learning rate parameter) was defined by empirical tests. 
Care must be taken in specifying the number of neurons. Too many can lead the NN 
to memorize the training data (over fitting), instead of extracting the general 
features that allow the generalization. Too few neurons may force the NN to 
spend too much time trying to find an optimal representation and thus wasting valuable 
computation time. The automatic configuration of NN is a recent topic of a discussion by
\cite{Sambatti12}.

One strategy used to collect data and to accelerate the processing of the MLP-NN 
training was to divide the entire globe into six regions: for the Northern Hemisphere, 
$90^o$ N and three longitudinal regions of  $120^o$ each; for the Southern Hemisphere, 
$90^o$ S and three longitudinal regions of  $120^o$ each. 
This division is based on the size of the regions, but the number of observations is distinct, as illustrated by Fig. 2. 
This regional division is applied only for the MLP-NN; the LETKF procedures are not modified. 

The MLP-NN were developed with a set of thirty NN (six regions with five prognostic 
variables ($p_s,$\textit{ u, v, T}, and \textit{q}). One MLP-NN with 
characteristics described above, was designed for each meteorological variable of the 
SPEEDY model and each region. Each MLP has two \textit{inputs} 
(model and observation vectors), one  \textit{output} neuron which is the 
\textit{analysis} vector, and eleven neurons in a hidden layer. 
The activation function used to ensure nonlinearity of the problem is the 
hyperbolic tangent (Eq.~\ref{tanh}), and learning rate parameter is defined to 
each variable set of networks, and the training scheme is the back-propagation algorithm. 
 
The MLP-NN is designed to emulate the LETKF. 
Fortran90 codes for SPEEDY and LETKF [originally developed by \cite{Miyoshi05}] were adapted to 
create the training dataset. The upper levels and the surface covariance error matrices to 
run the LETKF system, as well as the SPEEDY model boundary conditions data and physical 
parametrizations, are the same as those used for Miyoshi's experiments.

The initial process to run the implementation of the model assumes 
that it is perfect (initialization=0); and the SPEEDY model T30L7
integrated for one year of spin-up, i.e. the period required for a model 
to reach steady state and obtains the simulated atmosphere. 
The integration run for 01 January 1981 until 31 December 1981 and the result was the initial condition for SPEEDY 
to 01 January 1982, the date to initiate the experiment.

The output model fields, so-called ''true" model, were generated without 
assimilation (each 6-hours forecast of one execution is the initial condition for the next execution).
The ''true" (or control) model forecasts collected for model executions without observations,
considered four times per day (0000, 0600, 1200, 1800 UTC), 
from 01 January 1982 through 31 December 1984.

\begin{figure*}[t]
\vspace*{2mm}
  \begin{center}
  \noindent\includegraphics[width=12cm]{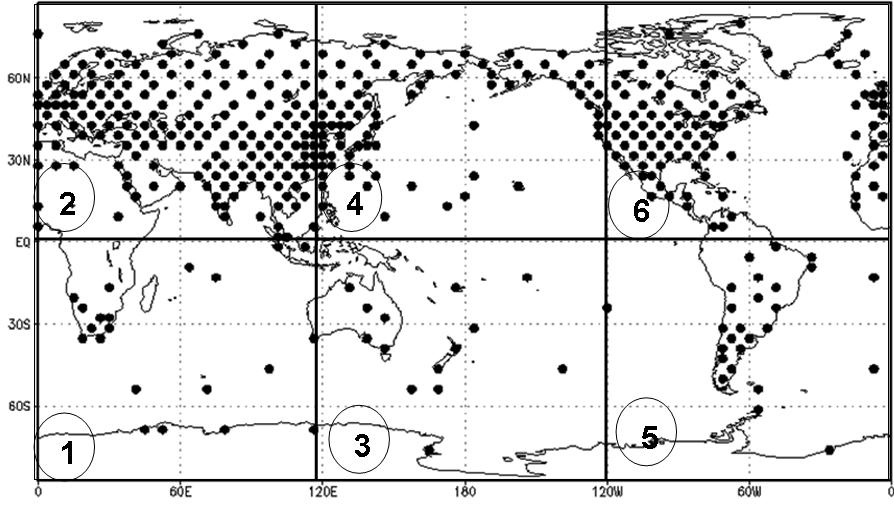}
    \end{center}
  \caption{Observations localization in global area.
  The dot points represent radiosonde stations (about 415) divided in six regions.}
  \label{fig:fig3}
\end{figure*}

The synthetic observations were generated, reading the true SPEEDY model fields, and 
adding a random noise of standard deviation error from meteorological variables: 
surface pressure ($p_s$), zonal wind component ($u$), vertical wind component ($v$), 
absolute temperature ($T$), and specific humidity ($q$) at each grid point where is located
an observations.  The variables were located at all grid points of the model. 
An observation \textit{mask} was designed, adding a positive flag at grid points 
where the observation should be considered, the locations chosen simulate the 
WMO data stations observations from radiosonde (see Fig. \ref{fig:fig3}). 
Except for $p_s$ observations, the other observations are upper level with
seven levels. Both assimilation schemes, LETKF and MLP-NN, use the 
same number of observations at the same grid point localization.

\vspace*{5mm}
\subsection{Training Process}
\noindent
The training process for the experiment is conducted with forecasts obtained from the 
SPEEDY model and LETKF analyses for the input/target vectors.
The LETKF analyses are executed with synthetic observations: upper-levels wind,
temperature and humidity, and surface pressure at 0000 and 1200 UTC (at 12,035 points), 
and only surface observations at 0600 and 1800 UTC (at 2,075 points). 
These LETKF processes generated the observations and analysis vectors,
and forecasts vectors for NN inputs. 
The analysis vector is the target for training the MPL-NN with back-propagation algorithm.
The analyses and forecasts are the average of the ensemble fields.

Executions of the model with the LETKF data assimilation were made for the 
same period mentioned for the ''true'' model: from 01 January 1982 
through 31 December 1984. The ensemble forecasts of LETKF has 30 members. 
The  ensemble average of the forecast and analyses fields, to this training 
process, are obtained by running SPEEDY model with the LETKF scheme.

These data are collected, initially, by dividing the globe into two regions  
(northern and southern hemispheres), but the computational cost was high 
because the training process took one day for the performance to converge. 
Next, the two regions were divided each into three regions, for a total of 
six regions.  When we divided the globe into six regions, 
the training, for a set of NN, took about minutes. The training 
time is important because if the NN lose 
knowledge of the system,  the re-training is necessary. 
Then, we use this division strategy to collect the thirty input vectors (observations, 
mean forecasts, and mean analyses) at chosen grid points by the observation 
mask (see above), during LETKF process.
The NN training process begin after collecting the input vectors for whole
period (three years).

The MLP-NN data assimilation scheme has no error covariance 
matrices to inform the spread of different observations. 
Therefore, to capture the influence of observations from the neighbouring 
region around a grid point, it is necessary to consider these observations. 
This calculation was based on the distance from the grid point related 
to observations inside a determined neighbourhood (initially:$\gamma = 0$)
\begin{eqnarray} \label{eq:M_pobs}
\hat{y}^o_{i \pm m,  j \pm m, k \pm m} =  
 \frac{y^o_{ijk}}{(6-\gamma)\, r^2_{ijk}} 
   + \sum_{l=1}^6 \alpha_l \frac{y^o_{i \pm m,  j \pm m, k \pm m}}{r^2_{ijk}}  \\
   (m = 1, 2, \ldots , M) \nonumber
\end{eqnarray}
\begin{equation}
\alpha_l = \left\{ \begin{array}{ll}
       0 & \mbox{(if there is no observation)}  \\ 
       1 & \mbox{(if there is observation, and:} \gamma_{\rm new} = \gamma_{\rm old}+1) \\
       \end{array}\right. \nonumber
\end{equation}
where $\hat{y}^o$ is the weighted observation,
 $M$ is the number of discrete layers considered around observation,
 $r_{ijk}^2 = (x_p-y^o_i)^2 + (y_p-y^o_j)^2+(z_p-y^o_k)^2$, where $(x_p, y_p, z_p)$ 
is coordinate of the grid point, and the $(y^o_i, y^o_j, y^o_k)$ is the coordinate 
of the observation, and $\gamma$ is a counter of grid points with observations 
around that grid point $(y^o_i, y^o_j, y^o_k)$. If $\gamma=6$ there is no influence 
to be considered.

Each influence observation is a new grid observation location, 
hereafter referred to as pseudo-observation, 
which adds values to the three input vectors to NN training process. 
Then, the grid points to be considered in MPL-NN analysis is greater than 
grid points considered to LETKF analysis, although these calculations 
are made without interference on LETKF system.

The back-propagation algorithm stops the training process using the criteria 
cited at item 6 in this section, after obtaining the best set of weights; it is a 
function of \textit{smallest error} between the MPL-NN analysis and the target analysis,
(e. g. when the root mean square error between the 
calculated output and the input desired vector is less than $10^{-5}$) for all NN. 
The training process was carried out for one or two epochs for each NN.
The training for a set of 30 NN took about fifteen minutes  
to get the fixes weights before the MLP-NN data assimilation cycle or
generalization process of MLP-NN.
\vspace*{5mm}

%
%
\subsection{Generalization Process}
\noindent
The training was performed with combined data from January, February, and 
March of 1982, 1983, and 1984 and MLP-NN is able to perform 
an analysis similar to the LETKF analysis in generalization process.

The generalization process is indeed, the data assimilation process. 
The MLP-NN results a analysis field. The MLP-NN activation was entering 
by input values at each grid point once, with no data used in the training
process. The input vectors are done at grid point where is marked with observation 
or pseudo-observation (Eq.~\ref{eq:M_pobs}).  
The procedure was the same for all NN, but one NN for each region 
and each prognostic variable, has different connection weights. 
The set of NN has one hidden layer, with the same number of neurons for all regions. 
 The grid points are put in the global domain to make the 
 analysis field after generalization process of MLP-NN, e.g. 
the activation of the set of 30 NN results a global analysis.
The regional division is only for inputting each NN activation.
 
The MLP-NN data assimilation was performed for one-month cycle. 
It started at 0000 UTC 01 January 1985, generating the 
initial condition to SPEEDY model and running the model to get 6-hours forecast 
for the next execution, i.e., the 0000 UTC cycle runs MLP-NN with 
observations for 0000 UTC of 01 January 1985 and 
the 6-hours forecast from 1800 UTC of 31 December 1984l, 
the result of MLP-NN is the initial condition of 0000 UTC 
to run the SPEEDY model and its result is the forecast to 0006 UTC, 
which is used to the 0006 UTC cycle, and so on. 
 
In this experiment the MLP-NN begins the activation in 01 January 1985 
and generates analyses and 6 hours forecasts up through 31 January 1985. 
\vspace*{5mm}

%
\section{Results}
\noindent
 The input and output values of prognostic variables ($p_s$, 
\textit{u}, \textit{v}, \textit{T}, and \textit{q}) are processed 
on grid points for time integrations to an intermittent forecasting 
and analysis cycle. Two discrete layers $M=2$ around a given 
observation are considered: in Eq.~\ref{eq:M_pobs}.

\begin{figure*}[t!]
\begin{center}
\hspace*{-0.5in}
\includegraphics[width=6.15in]{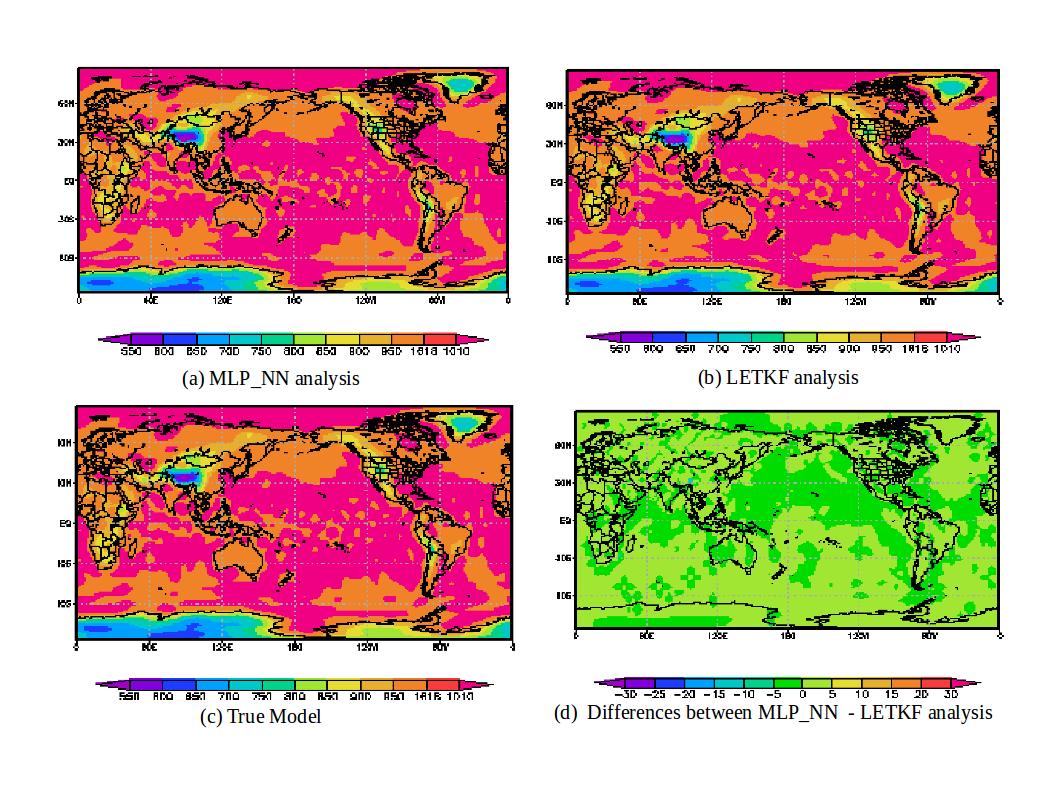}
\vspace*{-0.5in}
\caption{Surface Pressure (PS) [Pa] Fields - Jan/11/1985 at 18 UTC}
\end{center}
\label{fig:pres}
\end{figure*}

\begin{figure*}[t!]
\begin{center}
\hspace*{-0.5in}
\includegraphics[width=6.15in]{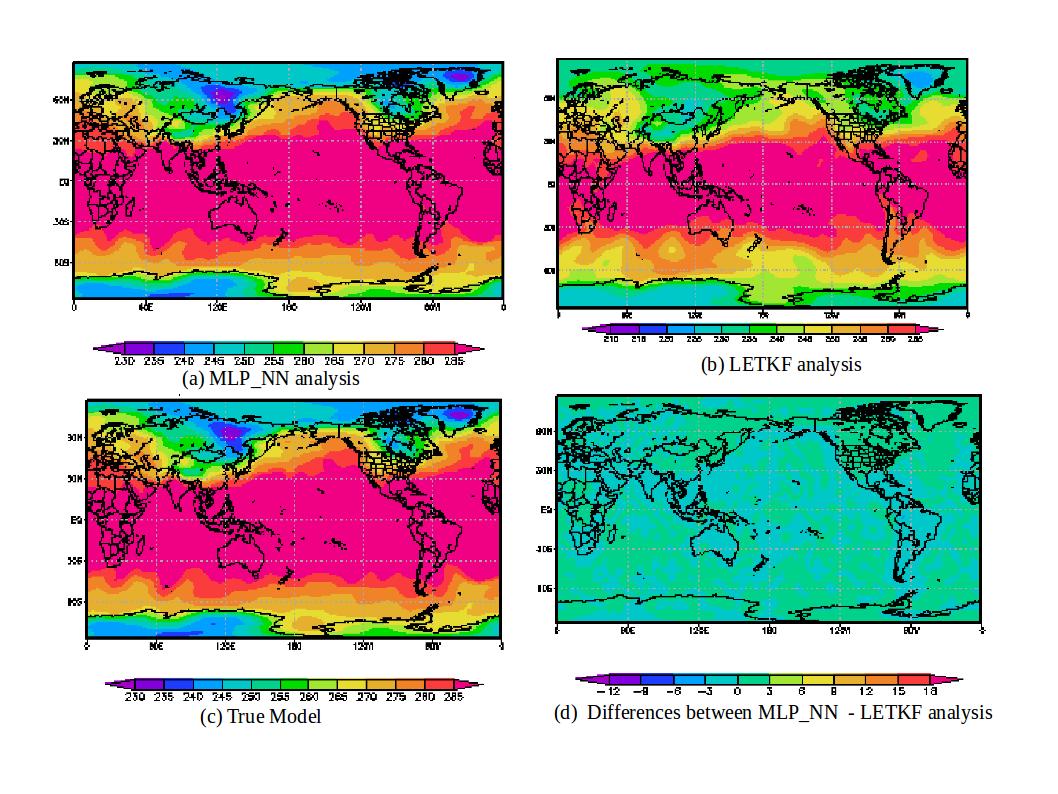}
\vspace*{-0.5in}
\caption{Absolute Temperature (T) [K] Fields at 950 hPa - Jan/11/1985 at 18 UTC}
\end{center}
\label{fig:res2}
\end{figure*}

\begin{figure*}[t!]
\begin{center}
\hspace*{-0.5in}
\includegraphics[width=6.1in]{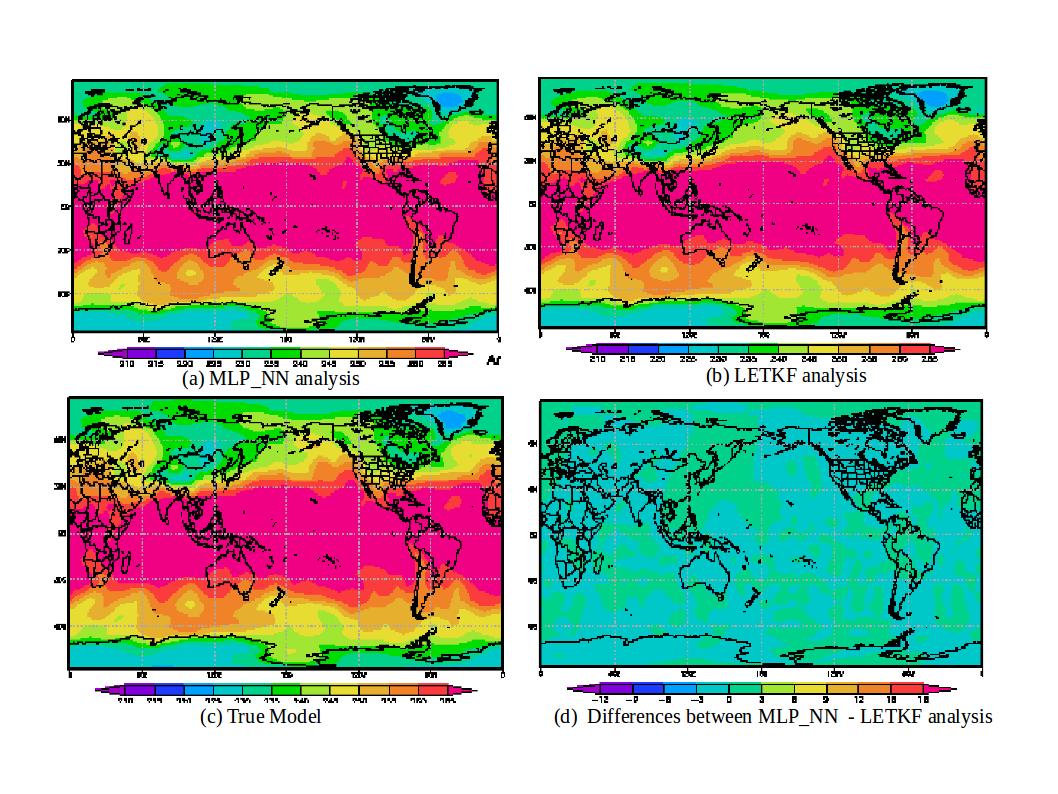}
\vspace*{-0.5in}
\caption{Absolute Temperature (T) [K] Fields at 500 hPa - Jan/11/1985 at 18 UTC}
\end{center}
\label{fig:res3}
\end{figure*}

\begin{figure*}[t!]
\begin{center}
\hspace*{-0.5in}
\includegraphics[width=6.1in]{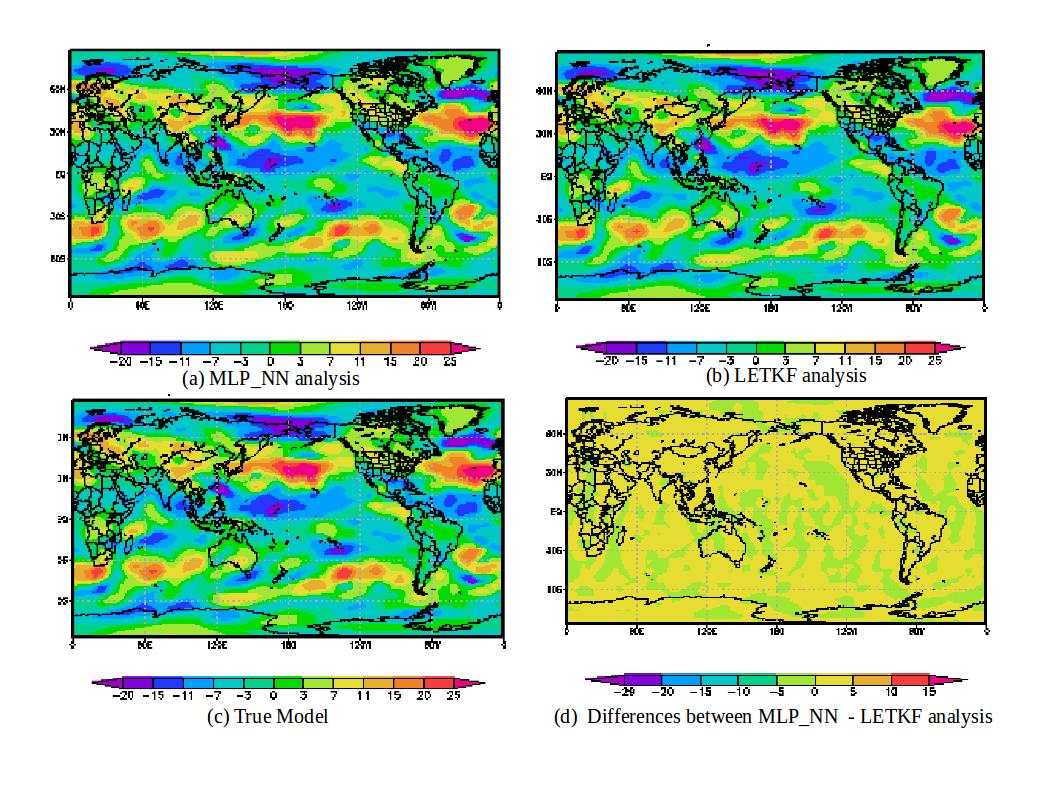}
\vspace*{-0.5in}
 \caption{Zonal Wind Component (u) [m/s] Fields at 950 hPa - Jan/11/1985 at 18 UTC}
 \end{center}
\label{fig:res4}
\end{figure*}

\begin{figure*}[t!]
\begin{center}
\hspace*{-0.5in}
\includegraphics[width=6.1in]{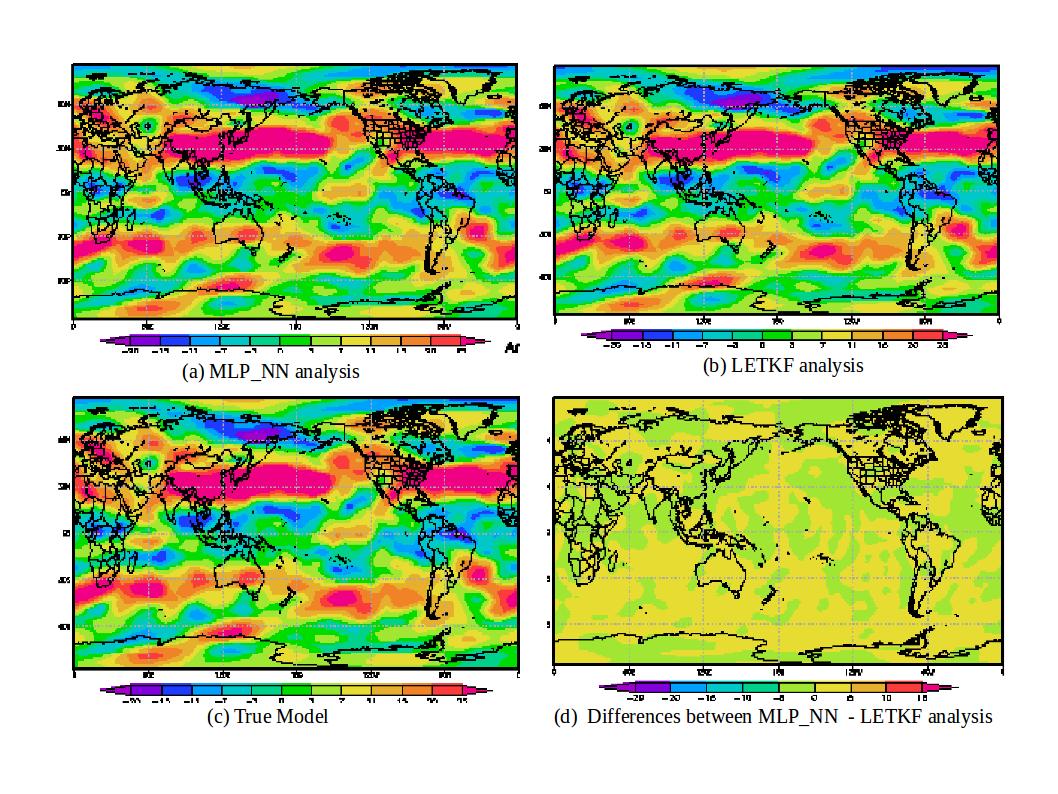}
\vspace*{-0.5in}
\caption{Zonal Wind Component (u) [m/s] Fields at 500 hPa - Jan/11/1985 at 18 UTC}
\end{center}
\label{fig:res5}
\end{figure*}

\begin{figure*}[t!]
\begin{center}
\hspace*{-0.5in}
\includegraphics[width=6.1in]{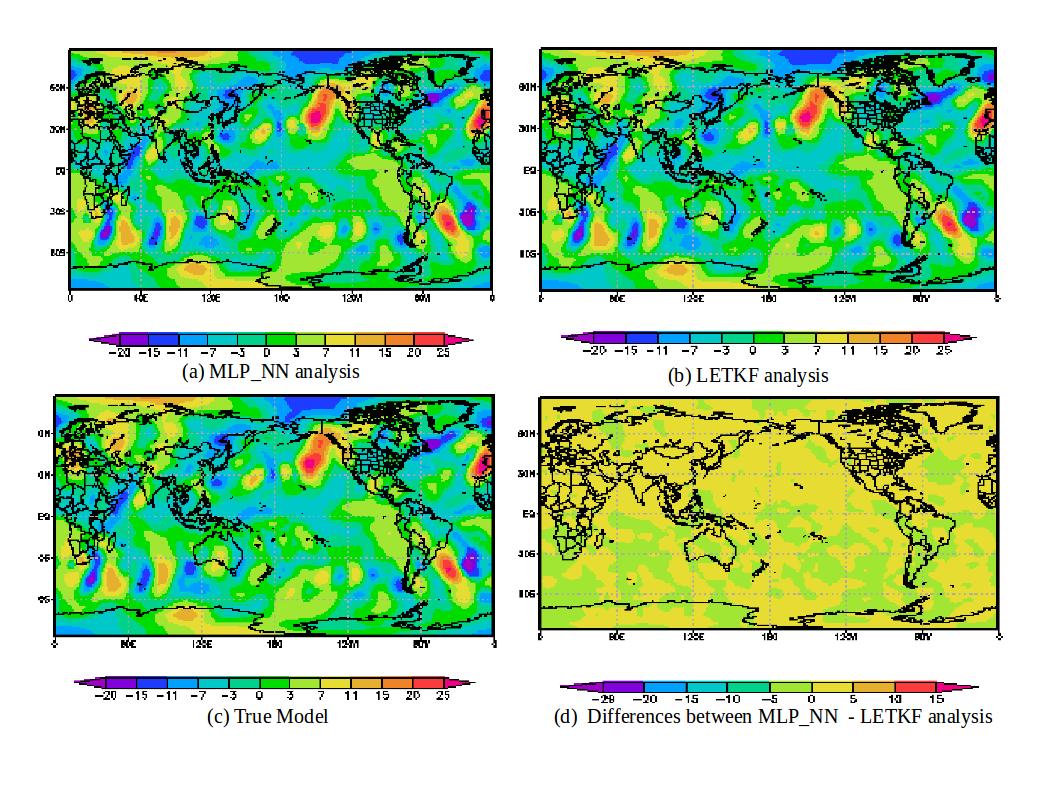}
\vspace*{-0.5in}
\caption{Meridional Wind Component (v) [m/s] Fields at 950 hPa - Jan/11/1985 at 18 UTC}
\end{center}
\label{fig:res6}
\end{figure*}

\begin{figure*}[t!]
\begin{center}
\hspace*{-0.5in}
\includegraphics[width=6in]{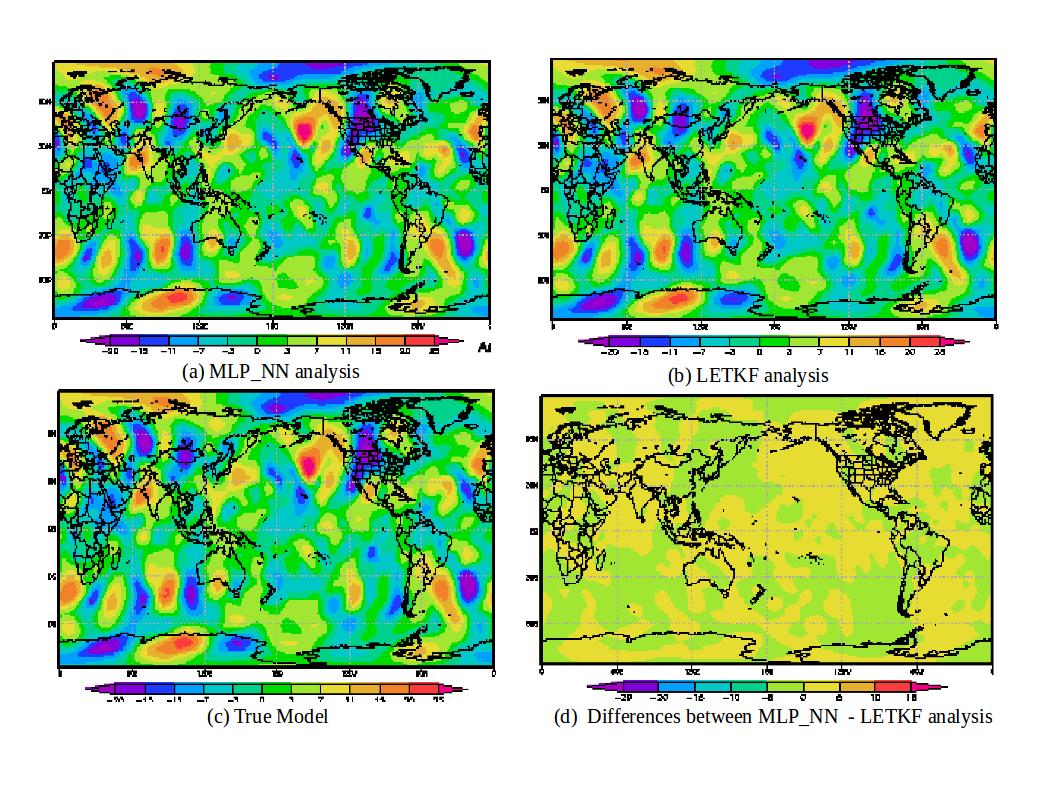}
\vspace*{-0.5in}
\caption{Meridional Wind Component (v) [m/s] Fields at 500 hPa - Jan/11/1985 at 18 UTC}
\end{center}
\label{fig:res7}
\end{figure*}

\begin{figure*}[t!]
\begin{center}
\hspace*{-0.5in}
\includegraphics[width=6.2in]{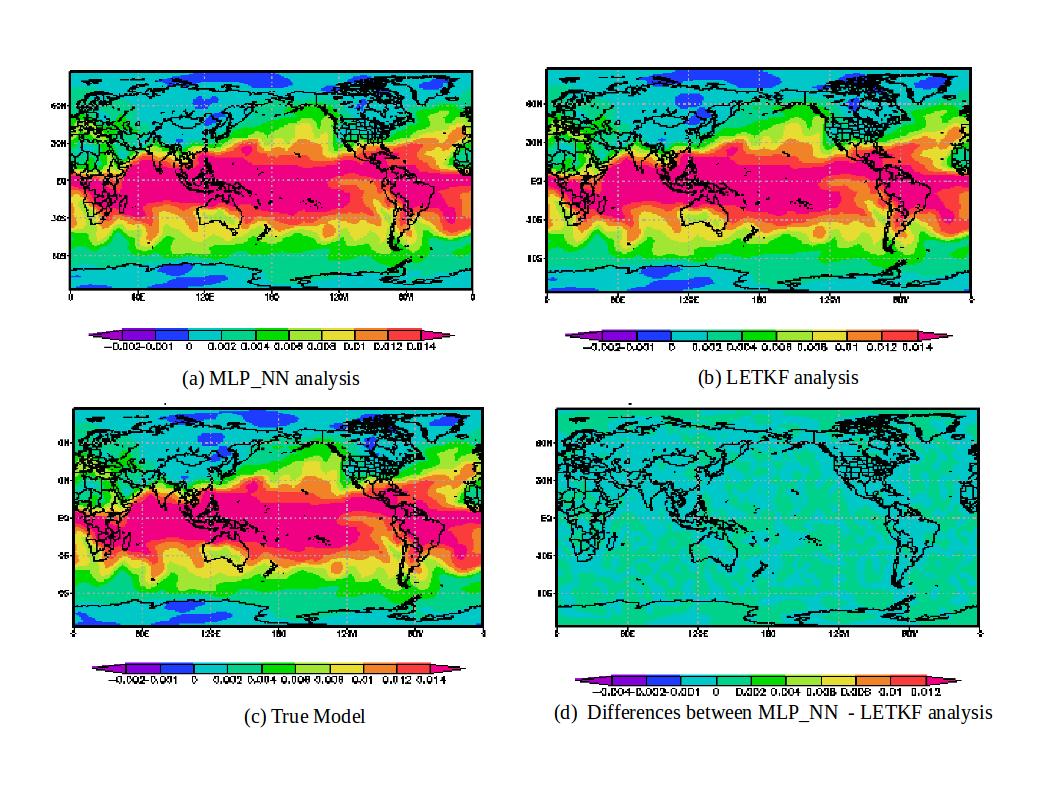}
\vspace*{-0.5in}
\caption{Specific Humidity (q) [kg/kg] Fields at 950 hPa - Jan/11/1985 at 18 UTC}
\end{center}
\label{fig:res8}
\end{figure*}

\begin{figure*}[t!]
\begin{center}
\hspace*{-0.5in}
\includegraphics[width=6.1in]{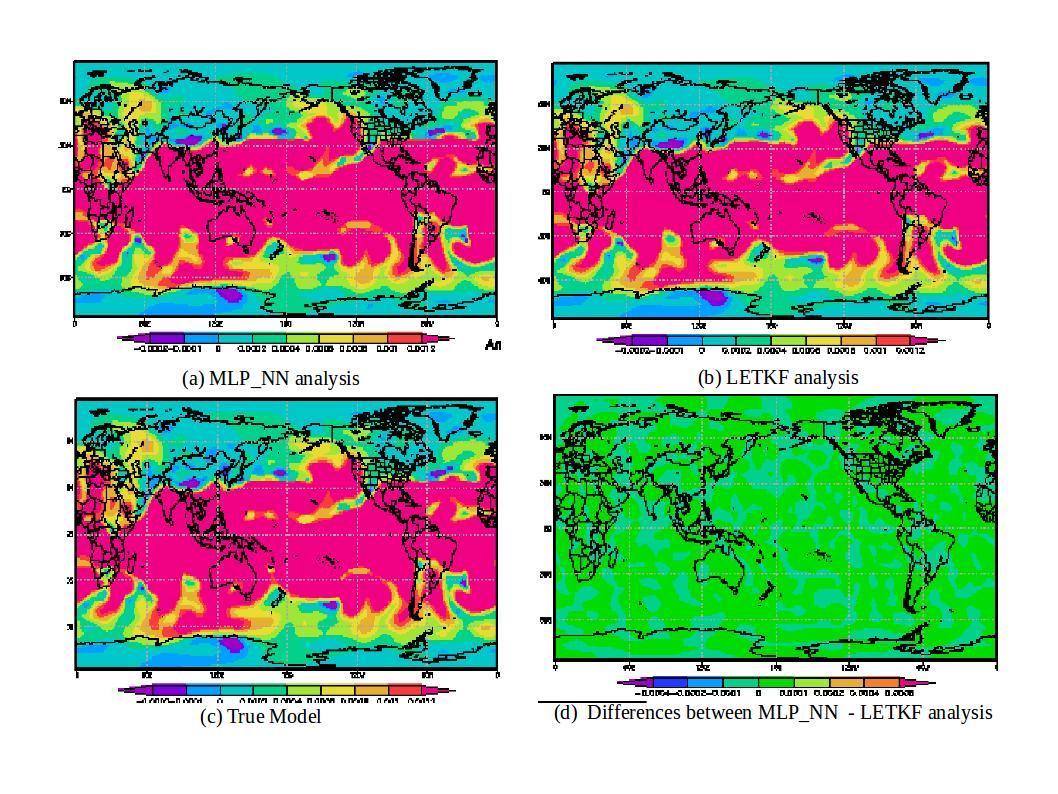}
\vspace*{-0.5in}
 \caption{Specific Humidity (q) [kg/kg] Fields at 500 hPa - Jan/11/1985 at 18 UTC}
 \end{center}
\label{fig:res9}
\end{figure*}

The results show the comparison of analyses fields, 
generated by the MLP-NN and the LETKF data assimilation schemes,
 and the true model field. 
 The global surface pressure fields (at 11 January 1985 at 1800  UTC) 
and differences between the analyses are shown in Fig. \ref{fig:pres}.
The analyses fields and the differences between both, for 11 January 1985 at 
1800  UTC at 950 hPa and 500 hPa, are also shown, 
for \textit{T, u, v} and \textit {q} meteorological global fields, 
in Figs.~\ref{fig:res2} - \ref{fig:res9}.
These results show that the application of MLP-NN, as an assimilation system, 
generates analyses similar to the analyses calculated by the LETKF system. 
Sub-figure (d) of Figs.~\ref{fig:res2} - \ref{fig:res9}
shows that the differences between the MLP-NN and LETKF analyses are very small,
we can verify the difference field of absolute temperature at 500 hPa, 
the differences are about  $3^o$ degrees;
or the differences field of humidity at 950 hPa, are about $0.002$.





\begin{figure*}[t!]
\vspace*{2mm}
\begin{center}
\hspace*{-0.5in}
\includegraphics[width=5.2in]{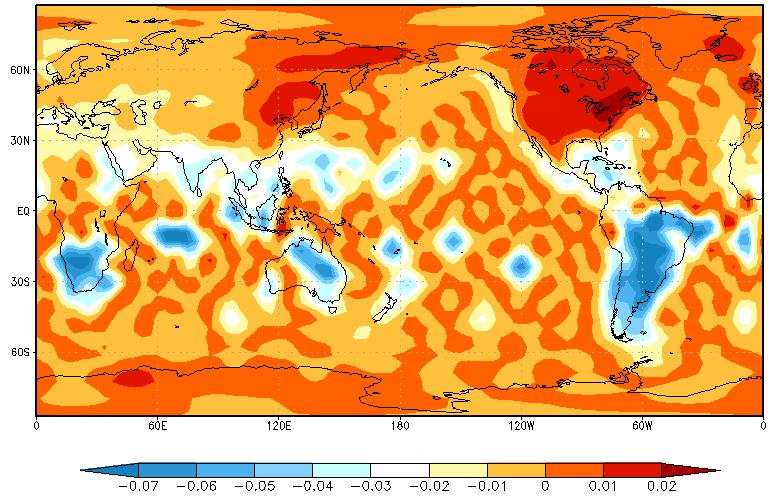}
\caption{Differences field of the average of absolute temperature MLP-NN analysis and LETKF 
             analysis for the assimilation cycle.}
\end{center}
\label{fig:average}
\end{figure*}

Monthly mean of absolute temperature analyses fields were obtained; the differences
field between the analyses (LETKF and MLP-NN) for data assimilation cycles 
is shown in Fig. \ref{fig:average}.
The differences are slightly larger in some regions, such as the northeast regions of 
North America and South America.

\begin{figure*}[t!]
\vspace*{2mm}
\begin{center}
\includegraphics[width=6.2in]{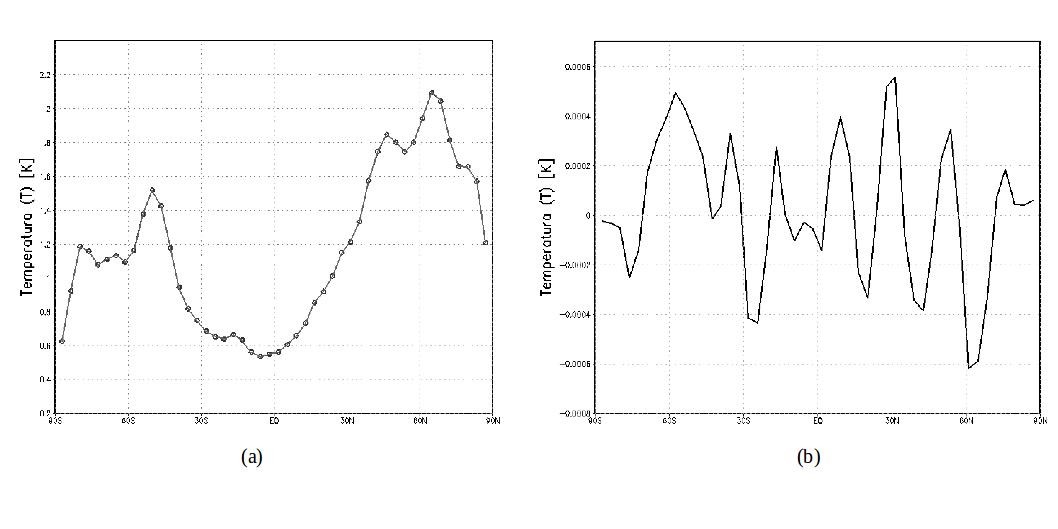}
\vspace*{-0.5in}
  \caption{Meridional Root Mean Square Error for entire period of the 
   assimilation cycles. RMSE analyses to the ''true'' state, 
   (a) the errors of the LETKF analysis (line) and 
   the errors of MLP-NN analysis (circles) to the absolute temperature at 500 hPa,  
   (b) the differences of RMSE analyses.}
  \label{fig:medt4}
  \end{center}
\end{figure*}

The root mean square error (RMSE) of the absolute temperature analyses to true model, 
are calculated by fixing a point in longitude ($87^o$W) for all latitude points.  
Fig.~\ref{fig:medt4} shows the temperature RMSEs for the entire period of the 
assimilation cycle (January 1985). Subfigure (a) for Fig.~\ref{fig:medt4}
shows the RMSE of the LETKF analysis by line, and the RMSE of the MLP-NN 
analysis by circles; and subfigure (b) for Fig. \ref{fig:medt4} shows the 
differences between LETKF and MLP-NN analyses RMSE. The differences 
are less than $10^{-3}$.
\vspace*{3mm}
%
\subsection{Computer Performance}
\noindent
 Several aspects of modeling stress computational systems and push capability requirements. 
 These aspects include increased resolution, the inclusion of improved physics processes 
 and concurrent execution of Earth-system components - that is, coupled models (ocean 
 circulation, and environmental prediction, for example). Often, real-time necessities 
 define capability requirements. In data assimilation, the computational requirements 
 become much more challenging. Observations from Earth-orbiting satellites in operational 
 numerical prediction models are used to improve weather forecasts. However, using this 
 amount of data increases the computational effort. As a result, there is a need for an 
 assimilation method able to compute the initial field for the numerical model in the 
 operational window-time to make a prediction. At present, most of the NWP centers find 
 it difficult to assimilate all the available data because of computational costs and 
 the cost of transferring huge amounts of data from the storage system to the main 
 computer memory.

\begin{figure*}[t]
\begin{center}
\includegraphics[width=5in]{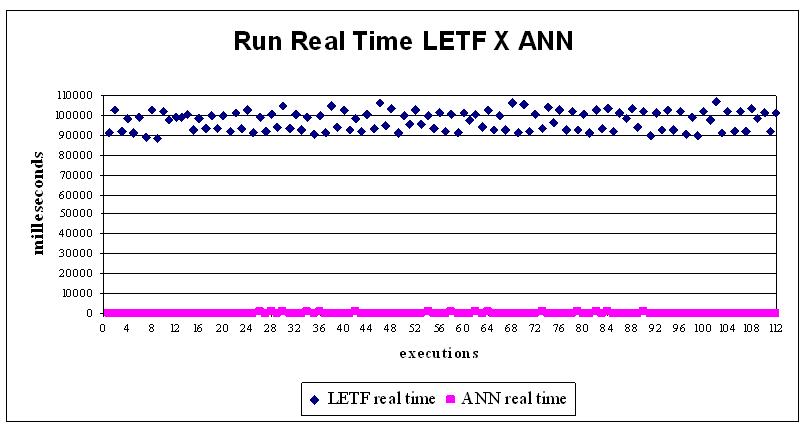}
\vspace*{ -0.1in}
  \caption{CPU-time for assimilation experiment: MLP-NN (magenta color), 
   and LETKF (blue color) methods.}
  \label{fig:perfor}
  \end{center}
\end{figure*}

 The data assimilation cycle has a recent forecast and the observations as the 
 inputs for  assimilation system. The latter MLP-NN system produced a analysis to initiate 
 the actual cycle, Fig.~\ref{fig:perfor} shows 112 cycles of data assimilation runs. 
 This time simulation experiment is for (28 days) January 1985. 
 There were 2,075 observations inserted at runs of 0600 and 
1800 UTC for surface variables and 12,035 observations inserted at runs of 0000 
and 1200 UTC for all upper layer variables.

The LETKF data assimilation cycle initiates, running the ensemble 
forecasts with the SPEEDY model and each analysis produced to each member 
at the latter LETKF cycle to result thirty 6-hours forecasts; 
the second step is to compute the average of those forecasts. 
After, with a set of observations and the mean forecast, the LETKF system is performed.
The LETKF cycle results one analysis to each member for the ensemble, and 
one average field of the ensemble analyses.
The MLP-NN data assimilation cycle is composed by the reading of 
6-hours. forecast of SPEEDY model from latter cycle and reading the set of observations
to the cycle time, the division of input vectors, the activation of MLP-NN and
the assembly of output vectors to a global analysis field.

The MLP-NN runtime measurement is magenta point of Fig.~\ref{fig:perfor}; 
it initiates after reading the 6-hours. forecast of SPEEDY model from latter cycle; and
the set of observations, it is the time of generalization 
MLP-NN with dividing forecasting/observations and with gathering global analysis.
The LETKF runtime measurement is the blue point of Fig.~\ref{fig:perfor};
it initiates after reading the mean 6-hours forecast of SPEEDY model 
and the set of observations. The LETKF time include the results of 
30 analyses and one mean ensemble analysis. 

The comparison in Table~\ref{tab:tempos1} is the data assimilation cycles 
for the same observations points and the same model resolution to the same
time simulations. LETKF and MLP-NN executions are performed independently.

Considering the total execution time of those 112 cycles simulated, 
the computational performance of the MLP-NN data assimilation, is better than 
that obtained with the LETKF approach. These results show that the computational 
efficiency of the NN for data assimilation to the SPEEDY model, for the adopted resolution, 
is 90 times faster and produces analyses of the same quality (see Table~\ref{tab:tempos1}).
Considering only the analyses execution time of those 112 data assimilation processes
simulated, the computational efficiency of MLP-NN is 421 times faster than
LETKF process. The Table~\ref{tab:tempos2} shows the mean execution time of
each element to one cycle of: the LETKF data assimilation 
method (ensemble forecast and analysis) and
the MLP-NN method (model forecast and analysis). 
The computational efficiency of one MLP-NN execution, 
keeps the relationship about speed-up, comparing with one LETKF execution.
(421 times faster).   Details for this experiment can be found in \cite{Cintra10}.

\begin{table*}[t!]
\caption{Total running time of 112 cycles of complete data assimilation 
(analysis and forecasting)}
\vskip4mm
\centering
\begin{tabular}{lcr}
\hline \hline
\textbf{Execution of 112 cycles} & MLP-NN (hour:min:sec)   & LETKF  (hour:min:sec) \\ [0.5ex] 
\hline \hline
Analisys time      &  00:00:25  &  03:14:55    \\
\hline
Ensemble time    &  00:00:00  &  01:05:44    \\
\hline
Single model time    &  00:02:28  &  00:00:00    \\
\hline
Total time           &  00:02:53  &  04:20:39    \\ \hline 
\hline
\end{tabular}
\label{tab:tempos1}
\end{table*}

\section{Conclusions}
\noindent
 In this study, we evaluated the efficiency of the MLP-NN in an atmospheric 
 data assimilation context. The MLP-NN is able to emulate systems for 
 data assimilation. For the present investigation, the MLP-NN approach is 
 used to emulate the LETKF approach, which is designed to improve the 
 computational performance of the standard EnKF. The another experiments
 with the same methodology can be found in \cite{Cintra12, Cintra13}.
 
The NN learned the whole process of the mathematical scheme 
of data assimilation through training with the LETKF.  The results for the 
MLP-NN analysis are very close to the results obtained from the LETKF data 
assimilation for initializing the SPEEDY model forecast, i.e., the analyses 
obtained with MLP-NN for the analysis field are similar to analyses computed 
by the LETKF, the differences between MLP-NN and LETKF analyses to surface
pressure fields in hPa, for example, is about (-5 to 5) hPa.
However, the computational performance of the set of thirty
NN is better.  The MLP-NN assimilation speed-up the LETKF scheme.

The application of the present NN data assimilation methodology is proposed 
at the Center for Weather Prediction and Climate Studies (Centro de Previs\~ao de 
Tempo e Estudos Clim\'aticos - CPTEC/INPE) with operational numerical global 
model and observations using nowadays. 
\vspace*{3mm}

\section{Acknowledgements}
The authors thank Dr. Takemasa Miyoshi and Prof. Dr. Eug\^enia Kalnay for 
providing computer routines for the SPEEDY model and the LETKF system. 
This paper is a contribution of the Brazilian National Institute of 
Science and Technology (INCT) for Climate Change funded by CNPq 
Grant Number 573797/2008-0 e FAPESP Grant Number
2008/57719-9. Author HFCV also thanks to the CNPq (Grant number 311147/2010-0).

 \end{multicols}

\end{document}